\documentclass[letterpaper]{article} 
\usepackage{aaai2027}  
\usepackage[hyphens]{url}  
\usepackage{graphicx} 
\usepackage{natbib}  
\usepackage{caption} 
\usepackage{algorithm}
\usepackage{algorithmic}

\usepackage{booktabs}
\usepackage{tabularx}
\usepackage{array}

\usepackage[utf8]{inputenc}
\usepackage{booktabs}   
\usepackage{makecell}   
\usepackage{graphicx}   
\usepackage{xspace}
\usepackage{float}

\usepackage{amssymb}
\usepackage{amsmath}

\usepackage{arydshln}
\usepackage{xcolor}
\usepackage{subcaption}
\usepackage{multirow}
\usepackage{newfloat}
\usepackage{listings}
\DeclareCaptionStyle{ruled}{labelfont=normalfont,labelsep=colon,strut=off} 
\floatstyle{ruled}
\newfloat{listing}{tb}{lst}{}
\floatname{listing}{Listing}

\usepackage{booktabs}

\newcommand{\rooticon}{\raisebox{-1.0ex}{\includegraphics[height=3.0ex]{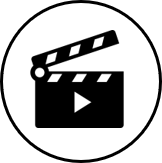}}}
\newcommand{\edgeicon}{\raisebox{-1.0ex}{\includegraphics[height=3.0ex]{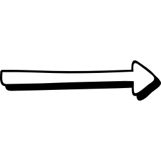}}}
\newcommand{\trajicon}{\raisebox{-1.0ex}{\includegraphics[height=3.0ex]{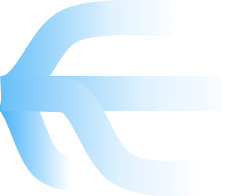}}}
\newcommand{\multishoticon}{\raisebox{-1.0ex}{\includegraphics[height=3.0ex]{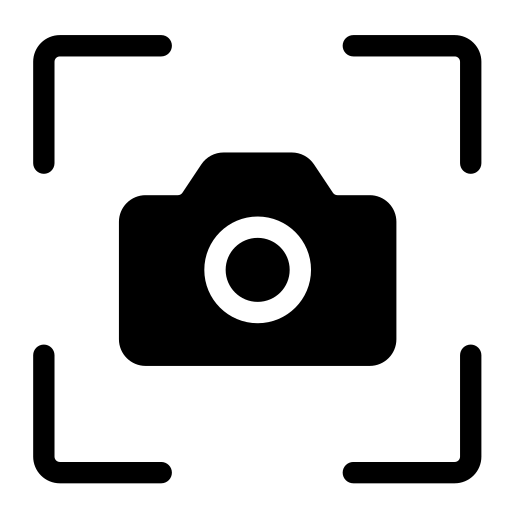}}}
\newcommand{\audioicon}{\raisebox{-1.0ex}{\includegraphics[height=3.0ex]{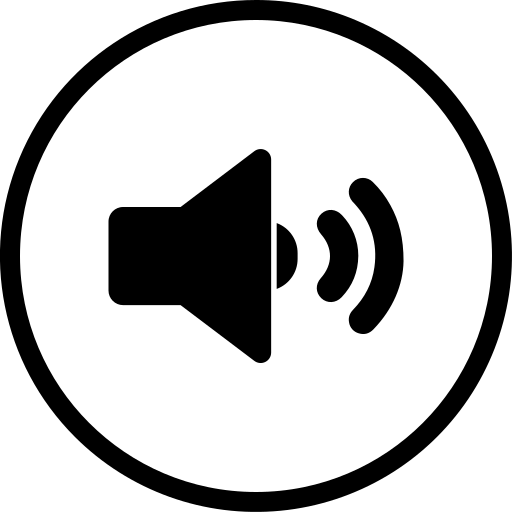}}}

\nocopyright

\title{ From Video Clips to Creation Trajectory: Sora100K for AI-Native Video Creation }

\author{
Sicong Yang\textsuperscript{\rm 1},
Ruihuan Yang\textsuperscript{\rm 1},
Jian Lu\textsuperscript{\rm 1},
Jianfei Yuan\textsuperscript{\rm 1},
Xiaodong Cun\textsuperscript{\rm 2},
Xiuli Bi\textsuperscript{\rm 1}
}

\affiliations{
\textsuperscript{\rm 1}Chongqing University of Posts and Telecommunications, Chongqing, China\\
\textsuperscript{\rm 2}GVC Lab, Great Bay University, Dongguan, China
}

\nocopyright

\makeatletter
\let\tempotmd@orig@maketitle\@maketitle

\def\@maketitle{%
  \tempotmd@orig@maketitle
  \par
  \begingroup
    \centering
    \noindent
    \makebox[\textwidth][c]{%
      \includegraphics[width=\textwidth]{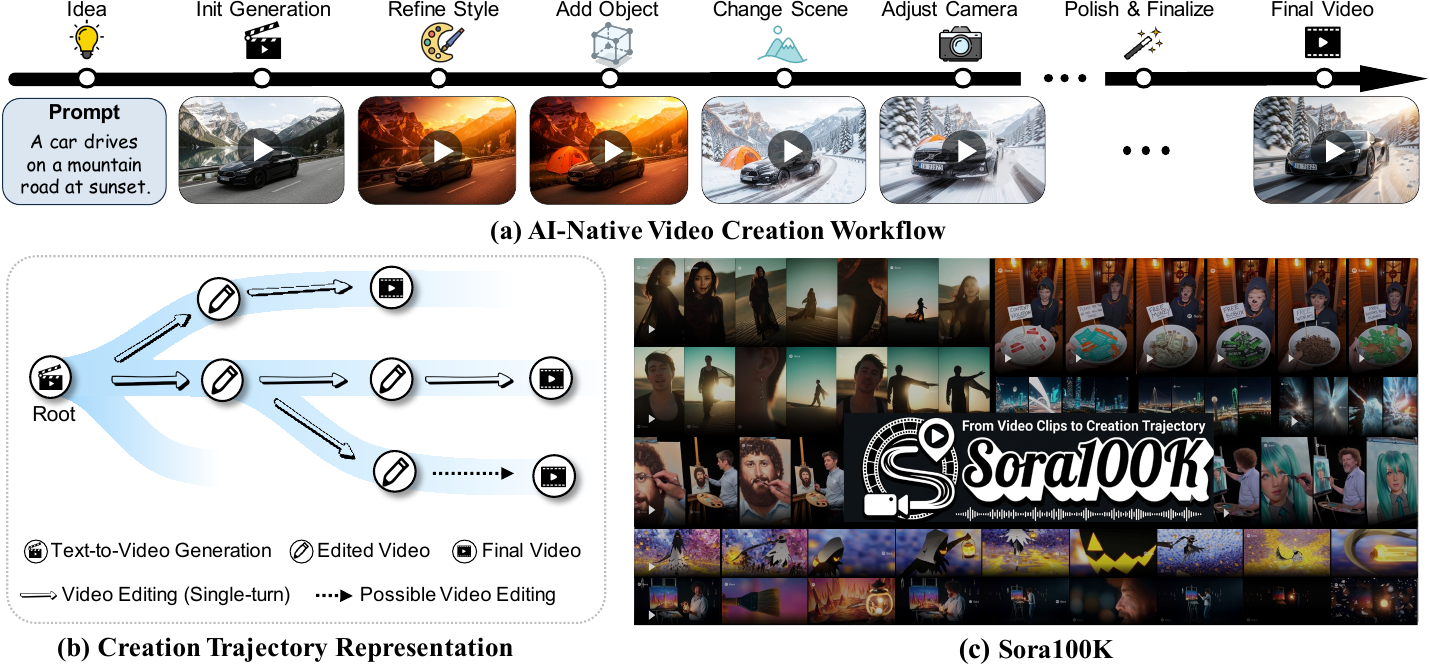}%
      }
    \par
    \captionof{figure}{Overview of Sora100K. We represent the AI-Native video creation workflow as a video creation trajectory that connects the generation root, successive editing edges, and a final video. Based on this representation, Sora100K organizes diverse video creation trajectories into a structured video dataset while preserving source-to-edit lineage and editing order.
    }
    \label{fig:teaser}
  \endgroup
  \par
}

\makeatother

\begin{document}

 \maketitle

\begin{abstract}

AI-Native video creation is shifting from isolated video clips toward iterative video creation workflows. However, existing datasets remain largely video clips, representing video generation and editing as separate tasks rather than connected stages of a video creation workflow. In this paper, we introduce Sora100K, a dataset that represents the AI-Native video creation workflow as a structured video creation trajectory. Specifically, we first identify video creation trajectories and decompose them into three subsets according to their structural roles: text-to-video generation records as roots, single-turn video editing records as editing edges, and multi-turn video editing records as complete trajectories. Then, we use a VLM to assign semantic annotations for generation roots and editing-operation annotations for editing edges. A strict construction pipeline further reconstructs source-to-edit lineage, editing order, and intermediate video states while ensuring data quality. Finally, we perform lightweight adaptation on LTX-2 models to assess the supervision value of Sora100K. The results show improvements in visual quality, multi-shot generation, and cross-shot consistency, while successive-turn evaluation reveals that following multi-turn editing instructions remains challenging. Sora100K establishes a new data foundation for AI-Native video creation beyond isolated video clips and toward structured video creation trajectory. The dataset and supplementary materials are publicly available at \url{https://huggingface.co/datasets/ysicong/Sora100K}.

\end{abstract}


\section{Introduction}
Commercial video generation models~\cite{openai2024sora, google2025veo31doc, runway2025gen4, kling2026kling30release, seedance2026seedance} continue to advance beyond short, isolated video clips toward cinematic videos with multi-shot structure and audio-visual synchronization. These advances are also changing how users create videos. Instead of relying on a single generation or editing step, users increasingly refine video content, visual style, camera motion, and audio through successive natural-language editing instructions. As illustrated in Figure~\ref{fig:teaser}, we represent this workflow from the initial prompt to the final edited video as a \emph{video creation trajectory}. It connects the initial generation with successive editing operations and intermediate video states.

However, as illustrated in Table~\ref{tab:structural_differences}, existing open academic datasets remained largely video clips. Text-to-video datasets~\cite{wang2025koala, chen2024panda} stored independent text--video pairs, while editing datasets~\cite{qin2024instructvid2vid, zi2025senorita} stored single-step source--instruction--target triplets. This organization disconnected initial generation from successive edits and omitted editing order and intermediate video states. The key gap therefore lies not only in scale, but also in data representation: isolated video clips do not capture connected video creation trajectory.

\begin{table}[!t]
\centering
\small
\setlength{\tabcolsep}{4pt}

\resizebox{\columnwidth}{!}{%
\begin{tabular}{l ccc cc}
\toprule
& \multicolumn{3}{c}{\textbf{Structure}} & \multicolumn{2}{c}{\textbf{Modality}} \\
\cmidrule(lr){2-4} \cmidrule(lr){5-6}
\textbf{Dataset}
& \textbf{Root}
& \textbf{Edge}
& \textbf{Trajectory}
& \textbf{Multi-shot}
& \textbf{Audio} \\
& \rooticon & \edgeicon & \trajicon & \multishoticon & \audioicon \\
\midrule
ActivityNet (CVPR’15)
& $\checkmark$ & $\times$ & $\times$ & $\times$ & $\checkmark$ \\

MSR-VTT (CVPR’16)
& $\checkmark$ & $\times$ & $\times$ & $\times$ & $\checkmark$ \\

WebVid-10M (ICCV’21)
& $\checkmark$ & $\times$ & $\times$ & $\times$ & $\times$ \\

HD-VILA-100M (CVPR’22)
& $\checkmark$ & $\times$ & $\times$ & $\checkmark$ & $\times$ \\

Panda-70M (CVPR’24)
& $\checkmark$ & $\times$ & $\times$ & $\times$ & $\times$ \\

Koala-36M (CVPR’25)
& $\checkmark$ & $\times$ & $\times$ & $\times$ & $\times$ \\

InstructVid2Vid (ICME’24)
& $\times$ & $\checkmark$ & $\times$ & $\times$ & $\times$ \\

VIVID-10M (arXiv’24)
& $\times$ & $\checkmark$ & $\times$ & $\times$ & $\times$ \\

Señorita-2M (NeurIPS’25)
& $\times$ & $\checkmark$ & $\times$ & $\times$ & $\times$ \\

OpenVE-3M (arXiv’25)
& $\times$ & $\checkmark$ & $\times$ & $\times$ & $\times$ \\

\midrule
\textbf{Sora100K (Ours)}
& $\checkmark$ & $\checkmark$ & $\checkmark$ & $\checkmark$ & $\checkmark$ \\
\bottomrule
\end{tabular}%
}

\caption{Structural and modality comparison of representative video datasets. Root and Edge denote text-to-video generation records and single-turn editing records, respectively, while Trajectory indicates whether they are connected through editing order and intermediate video states. Multi-shot and Audio specify the presence of multi-scene transitions and aligned audio tracks.}
\label{tab:structural_differences}
\end{table}

To bridge this gap, we introduce Sora100K, a large-scale dataset for video creation trajectory. As illustrated in Figure~\ref{fig:teaser}, it contains over 100K video records curated from the AI-Native social media platform Sora2~\cite{openai2024sora}  and organized into three structural subsets: text-to-video generation as generation roots, single-turn video editing as editing edges, and multi-turn video editing as complete video creation trajectories. These subsets jointly cover different stages of video creation trajectories. Beyond its trajectory-level organization, the proposed Sora100K explicitly preserves two properties that remain underexplored in open academic datasets, \emph{i.e.}, multi-shot structure and audio-visual synchronization. These characteristics make Sora100K closer to contemporary real-world video creation workflow. To make our dataset analyzable and reusable, we develop an annotation pipeline. We use Qwen2.5-VL-7B~\cite{bai2025qwen25vl} to scale the annotation of generation semantics and editing operations, these annotations provide structured information at the three subsets. We use manual inspection to verify the accuracy of the automated annotations. To evaluate the practical utility of Sora100K, we further apply lightweight LoRA~\cite{hu2022lora} adaptation to LTX-2~\cite{hacohen2026ltx}. The experimental results show that a small subset of Sora100K already provides effective supervision for video generation and editing.

In summary, our main contributions are as follows:
\begin{itemize}
\item We introduce a representation that structures AI-Native video creation as a creation trajectory with generation roots and editing edges.
\item We construct Sora100K, a large-scale dataset with semantic and editing annotations, source-to-edit lineage, intermediate video states, multi-shot structure, and audio-visual synchronization.
\item We provide structured annotations and trajectory-level organization that support
research on cross-turn dependencies, and maybe can support long-horizon video generation and editing in future research.
\end{itemize}

\section{Related Work}

\subsection{Text-to-Video Generation Models and Datasets.} 

Early video generation methods~\cite{vondrick2016generating,tulyakov2018mocogan}, including GAN-based and autoregressive approaches, mainly focused on unconditional or weakly conditioned generation, limiting alignment with complex text prompts. Diffusion-based models later improved text--video alignment through text encoders and large-scale text--image pretraining. Early video diffusion models~\cite{ho2022video,singer2022make} still struggled with temporal consistency, while later latent diffusion and spatiotemporal transformer architectures improved visual quality and temporal coherence. Recent models~\cite{seedance2026seedance,kling2026kling30release} further supported structured multi-shot generation and audio-visual synchronization.

Large-scale datasets such as WebVid-10M~\cite{bain2021frozen} and HD-VILA-100M~\cite{xue2022clip} provided the text--video pairs required to train foundational text-to-video models. Other datasets, including MSR-VTT~\cite{xu2016msr} and ActivityNet~\cite{caba2015activitynet}, were widely used for video understanding, captioning, and retrieval. These datasets generally organized samples as independent text--video pairs. In our workflow-centric representation, each pair corresponds to a generation root that can support subsequent editing operations. However, prior datasets did not connect generation roots with later editing edges and therefore did not preserve source-to-edit lineage, editing order, or intermediate video states.

\begin{figure*}[!t]
  \centering
  \includegraphics[width=\textwidth]{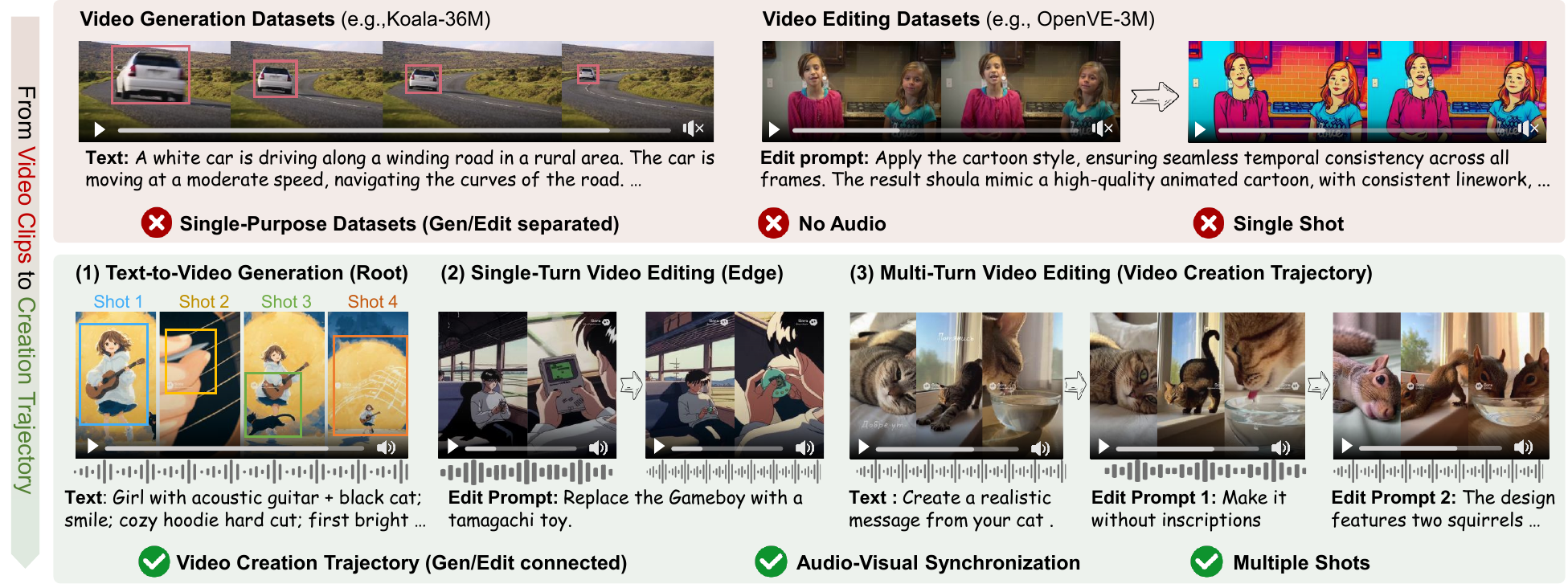}
  \caption{Comparison between representative video datasets and Sora100K. Prior datasets typically store video generation and editing as separate clip-level samples. Sora100K connects text-to-video generation roots, single-turn editing edges, and multi-turn editing trajectories, while preserving multi-shot structure and audio-visual synchronization.}
  \label{fig:diff}
\end{figure*}

\subsection{Video Editing Models and Datasets.} 

Instruction-driven visual editing first advanced in the image domain. InstructPix2Pix~\cite{brooks2023instructpix2pix} used model-generated paired data to learn general editing instructions. Extending this setting to video required temporal consistency across frames. Tune-A-Video~\cite{wu2023tune} and FateZero~\cite{qi2023fatezero} used one-shot tuning or attention mechanisms to preserve source structure, while TokenFlow~\cite{geyer2023tokenflow} and CoDeF~\cite{ouyang2023codef} modeled temporal correspondences for consistent editing. These methods mainly addressed individual editing operations, which correspond to editing edges in a video creation trajectory. Recent studies explored iterative editing and cross-turn consistency~\cite{lee2026memoryv2vmemoryaugmentedvideotovideodiffusion}, but data preserving successive prompts and videos remained limited.

Recent datasets expanded the scale and coverage of instruction-based video editing. InstructVid2Vid~\cite{qin2024instructvid2vid} constructed video--instruction triplets for general editing, while VIVID-10M~\cite{hu2024vivid} focused on large-scale local editing. Se\~norita-2M~\cite{zi2025senorita} and OpenVE-3M~\cite{he2025openve} further improved data quality and editing diversity. However, these datasets generally stored each operation as an independent source--instruction--target triplet. Such triplets provide supervision for individual editing edges but do not connect them to the generation root or preserve editing order and intermediate states.

\begin{figure*}[!t]
  \centering
  \includegraphics[width=\textwidth]{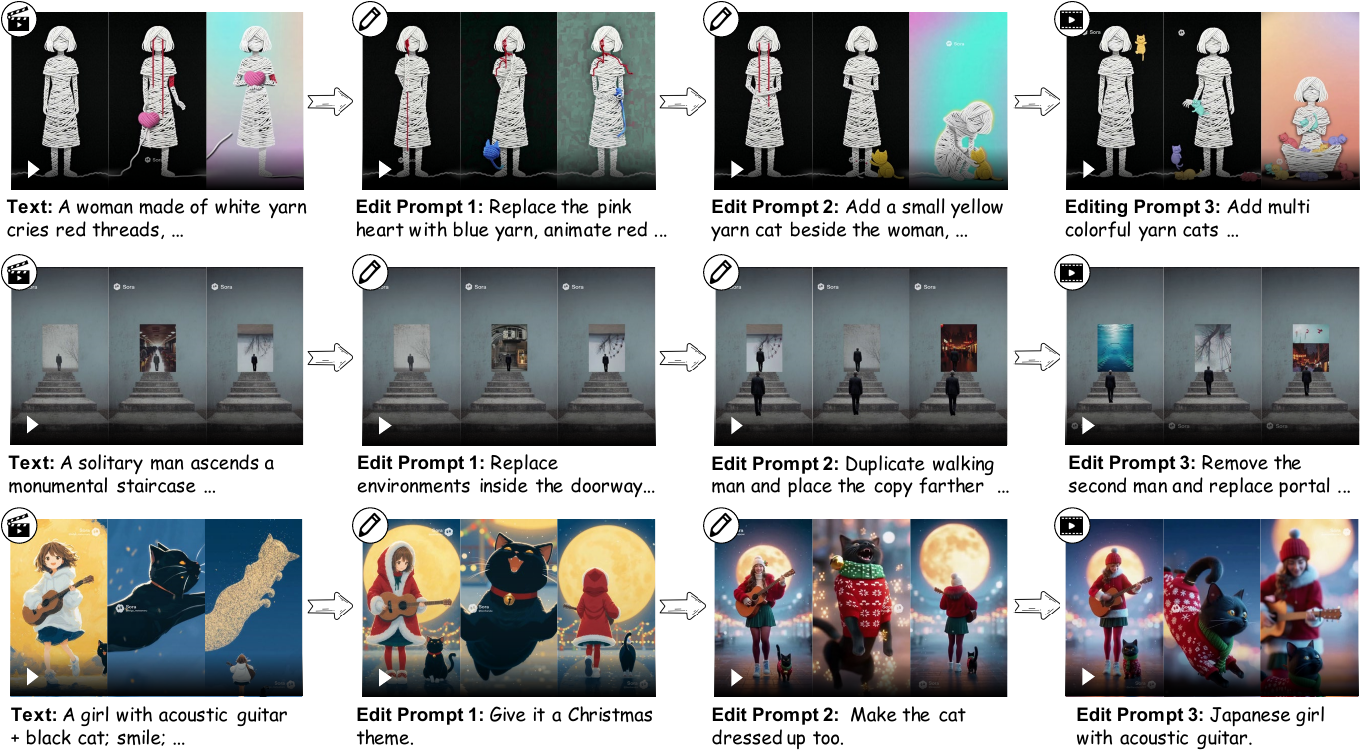}
  \caption{Sample video creation trajectories from Sora100K. In each row, the initial text-to-video generation record corresponds to the root. Each pair of adjacent video states and its editing instruction forms a single-turn editing edge, while the complete ordered sequence forms a video creation trajectory.}
  \label{fig:show}
\end{figure*}

\section{Dataset Characteristics}
In this section, we compared the differences between Sora100K and previous datasets. As shown in Table~\ref{tab:structural_differences} and Figure~\ref{fig:diff}, Sora100K represents AI-Native video creation as a structured creation trajectory that connects the video generation, video editing and creation trajectory, Sora100K moves beyond isolated video clips by modeling creation trajectories and retaining multi-shot and audio-visual information.

\subsection{Structure-Level Characteristics}

\hspace*{\parindent}\textbf{Video Creation Trajectories}.
AI-Native video creation begins with an initial generation and proceeds through successive editing operations. A creation tree represents all video states derived from the same initial prompt, while a creation trajectory is an ordered root-to-node path within the tree. 
\begin{equation}
\mathcal{C}_i =
\left(
p_i,\mathcal{V}_i,\mathcal{E}_i,r_i
\right),
\label{eq:creation_tree}
\end{equation}
where $\mathcal{C}_i$ denotes the $i$-th creation tree, $p_i$ denotes the initial generation prompt, $\mathcal{V}_i$ is the set of video states produced during the video creation trajectory, and $r_i=v_i^{0}\in\mathcal{V}_i$ is the root node corresponding to the initial generated video. As defined in Eq.~\ref{eq:creation_tree}, the tree connects the initial prompt, generated video states, and editing operations within one video creation trajectory. Its edge set is defined as
\begin{equation}
\mathcal{E}_i =
\left\{
\left(v_i^{a},e_i^{a\rightarrow b},v_i^{b}\right)
\right\},
\label{eq:editing_edges}
\end{equation}
where $v_i^{a}$ is the source video state, $e_i^{a\rightarrow b}$ is the editing instruction applied to it, and $v_i^{b}$ is the resulting video state.

A video creation trajectory connects the editing edges in Eq.~\ref{eq:editing_edges} into a root-to-node path in $\mathcal{C}_i$:
\begin{equation}
\mathcal{T}_{i,k} =
\left(
p_i,v_i^{0},
e_{i,k}^{1},v_{i,k}^{1},
\ldots,
e_{i,k}^{L_{i,k}},v_{i,k}^{L_{i,k}}
\right),
\label{eq:creation_trajectory}
\end{equation}
where $\mathcal{T}_{i,k}$ denotes the $k$-th trajectory derived from $\mathcal{C}_i$, $e_{i,k}^{t}$ and $v_{i,k}^{t}$ denote the editing instruction and resulting video at turn $t$, respectively, and $L_{i,k}$ is the total number of editing turns. When no branching occurs, the creation tree degenerates into a single trajectory as defined in Eq.~\ref{eq:creation_trajectory}.

The three subsets of Sora100K are not independent task collections. Instead, they provide three structural views of the creation trees defined in Eq.~\ref{eq:creation_tree}. Specifically, the text-to-video generation subset records the initial prompt and root node, the single-turn video editing subset records individual editing edges, and the multi-turn video editing subset records complete root-to-node trajectories:
\begin{equation}
\begin{aligned}
\mathcal{D}_{\mathrm{gen}}
&=
\left\{
\left(p_i,r_i\right)
\right\}_{i=1}^{N}, \\
\mathcal{D}_{\mathrm{edit}}
&=
\bigcup_{i=1}^{N}\mathcal{E}_i, \\
\mathcal{D}_{\mathrm{traj}}
&=
\left\{
\mathcal{T}_{i,k}
\right\}_{i,k},
\end{aligned}
\label{eq:unified_structure}
\end{equation}
where $N$ is the number of creation trees, and $\mathcal{D}_{\mathrm{gen}}$, $\mathcal{D}_{\mathrm{edit}}$, and $\mathcal{D}_{\mathrm{traj}}$ denote the text-to-video generation, single-turn video editing, and multi-turn video editing subsets, respectively. As summarized in Eq.~\ref{eq:unified_structure}, generation serves as the trajectory root, each single-turn edit forms a transition edge, and each multi-turn sequence constitutes a complete trajectory.


\subsection{Video-Level Characteristics}

Sora100K exhibits two properties: audio-visual synchronization and multi-shot structure, which were weakly supported in prior video generation and editing datasets. 

\textbf{Audio-Visual Synchronization}. 
Sora100K preserves audio signals that are closely aligned with the visual content. Earlier video datasets often contained visual-only samples because of the limited audio-generation capabilities of previous models, restricting their use for joint audio-visual modeling. In contrast, Sora100K retains the original synchronized audio without excessive processing, providing a practical data foundation for audio-visual video generation.

\textbf{Multi-Shot Structure}. 
Sora100K contains a large proportion of videos with clear multi-shot organization. Earlier datasets were largely dominated by single-shot clips and provided limited scene transitions and shot changes. Although the Sora2 technical report did not explicitly emphasize multi-shot generation, our analysis shows that most collected videos contain multiple shots and generally maintain character consistency across shot transitions.

\section{Dataset Construction and Annotation}
In this section, we describe how Sora100K is constructed, how its three subsets are annotated, and how each subset is composed within a unified and systematic framework. As illustrated in Figure~\ref{fig:show}, we present representative samples here and provide additional examples in the appendix.

\subsection{Data Curation and Quality Filtering}

We construct Sora100K from approximately 180K public Sora2~\cite{openai2024sora} records. After recovering prompts, editing instructions, and available source references, we apply accessibility, VBench-based quality, and traceability filtering, retaining approximately 137K, 111K, and finally 103.4K records.
We organize the filtered records according to their recovered source-to-edit relations. Every valid initial generation is retained as a root in the text-to-video generation subset, even when no subsequent edit is available. Every valid source--instruction--target relation is retained as an edge in the single-turn video editing subset. Records containing only one edit therefore contribute both a generation root and an editing edge, while only connected paths with at least three video states are included in the multi-turn video editing subset. This process yields 18,451 generation roots, 76,964 editing edges, and 8,024 multi-turn trajectories.
And we inspect representative trajectories and find their source-to-edit lineage, editing order, and intermediate states to be consistent with the recovered metadata. We use Qwen2.5-VL-7B~\cite{bai2025qwen25vl} to annotate root semantics and edge-level editing operations, and further verify the automatic annotations through human evaluation.



\begin{figure*}[t]
    \centering
    \begin{subfigure}[b]{0.36\textwidth}
        \centering
        \includegraphics[width=\textwidth]{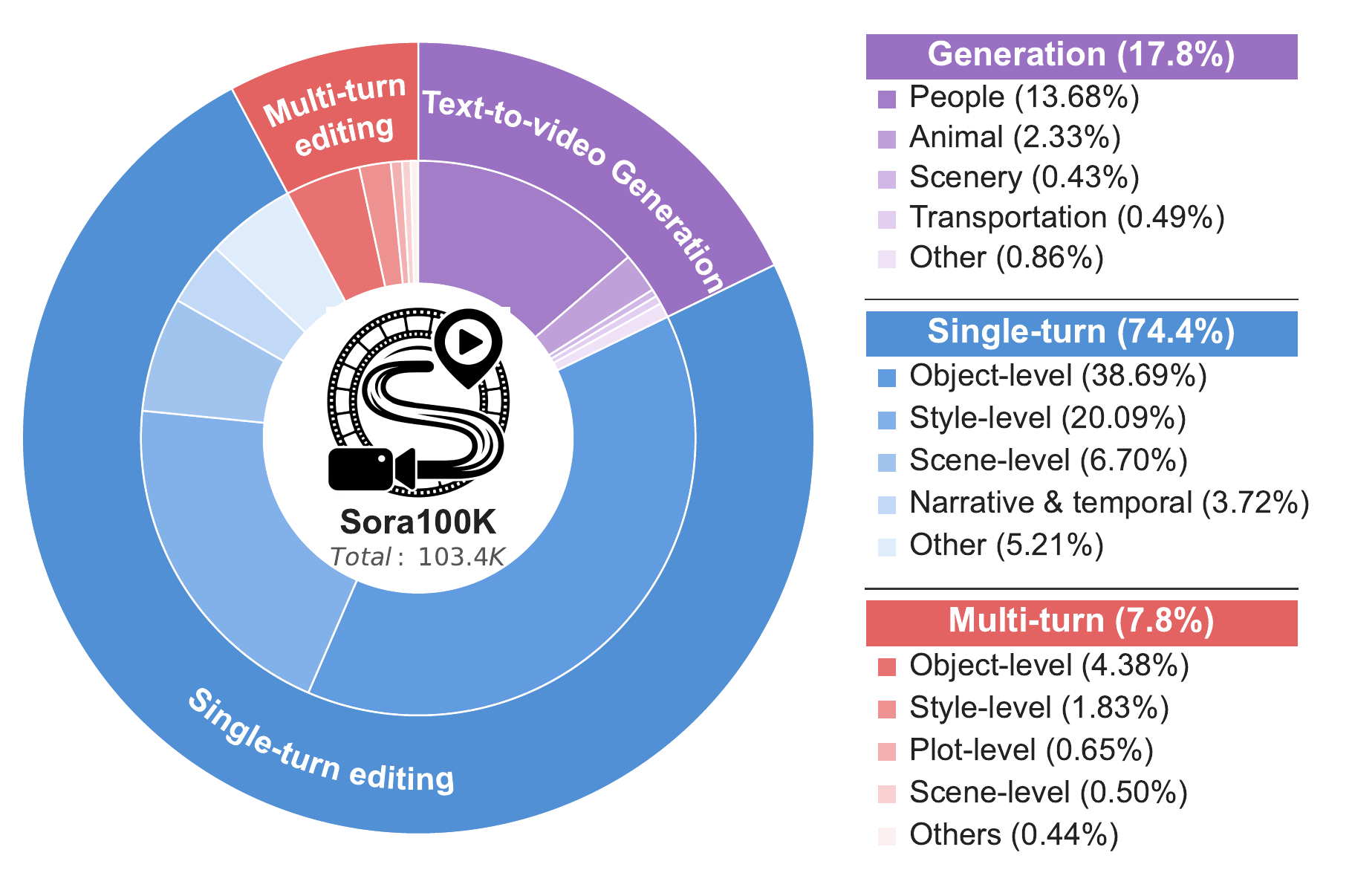}
        \caption{Overall composition}
        \label{fig:sub1}
    \end{subfigure}
    \hfill
    \begin{subfigure}[b]{0.62\textwidth}
        \centering
        \includegraphics[width=\textwidth]{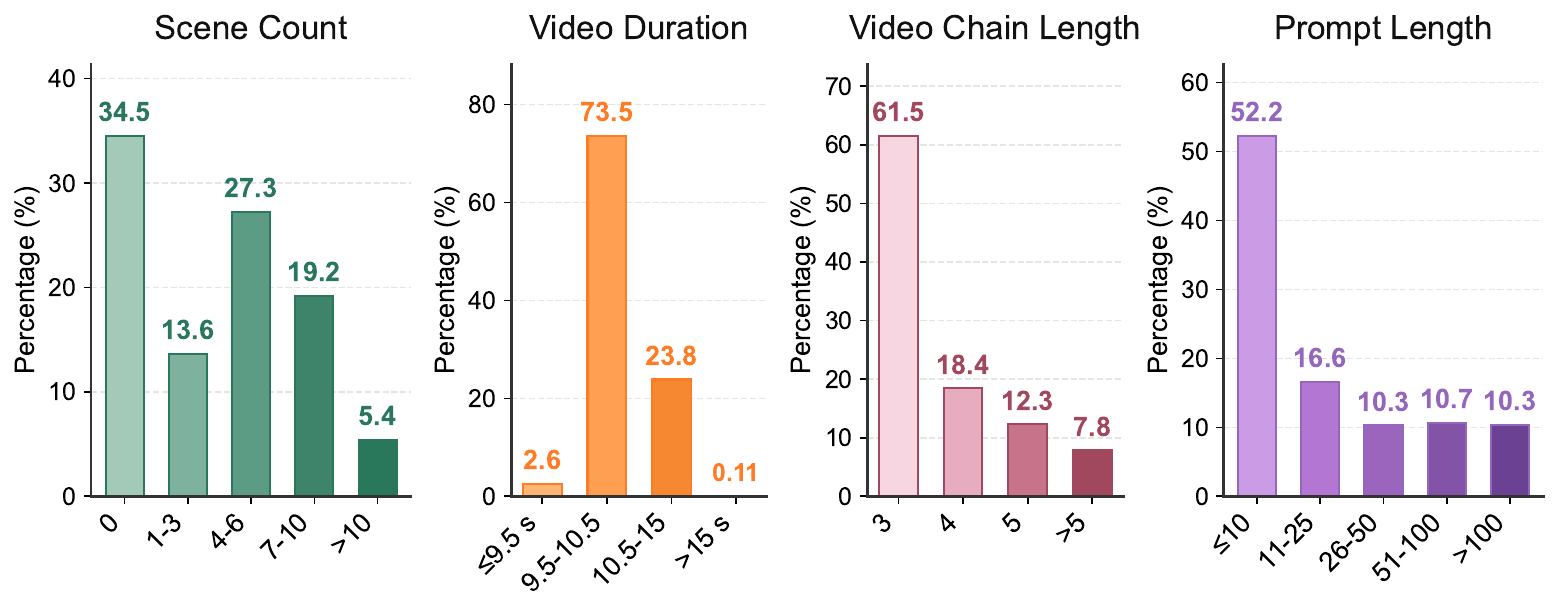}
        \caption{Video statistics and generation prompt length}
        \label{fig:sub4}
    \end{subfigure}
    \caption{Statistical analysis of Sora100K, including overall task proportions, video-level statistics, editing category distribution, generation category distribution, and prompt length distribution.}
    \label{fig:four_images_row}
\end{figure*}

\subsection{Dataset Annotation}
A video creation trajectory contains information at different structural levels. We therefore adopt a hierarchical annotation scheme over the text-to-video generation, single-turn video editing, and multi-turn video editing subsets. These annotations characterize the generation roots, editing operations, and the video creation trajectories, respectively.

\textbf{Root-Level Semantic Annotation.}
For the text-to-video generation subset, we characterize each generation root along five dimensions: semantic category, video duration, frame count, number of shots, and prompt length. We construct a single-label semantic taxonomy, and use Qwen2.5-VL-7B~\cite{bai2025qwen25vl} to annotate the samples. Prompt length is extracted from the original metadata, while the video-level attributes are obtained through video analysis. 

\textbf{Edge-Level Operation Annotation.}
For the single-turn video editing subset, we characterize each editing edge by its editing category, video duration, frame count, and number of shots. We construct a multi-label taxonomy because one edited may involve multi-modifications. We use Qwen2.5-VL-7B~\cite{bai2025qwen25vl} to annotate the samples. 

\textbf{Trajectory-Level Structural Annotation.}
For the multi-turn video editing subset, we preserve the editing category, source-to-edit lineage, and chain length of each video creation trajectory. These structural annotations connect the generation roots and editing edges into ordered creation trajectory and support the analysis of multi-turn dependencies. This structural information connects the annotated generation roots and editing edges into complete trajectories.

\subsection{Dataset Statistics}

We first examine the multi-turn video editing subset to characterize the trajectory-level structure of Sora100K, and followed by the generation roots and editing edges. The detailed distributions are shown in Figure~\ref{fig:four_images_row}. We present further statistical details and sample illustrations in the appendix.

\textbf{Trajectory-Level Structural Statistics.}
Sora100K has 8,024 multi-turn video editing trajectories, accounting for 7.8\% of its 103.4K structured records. As shown in Figure~\ref{fig:four_images_row}, we analyze editing-operation types rather than initial-generation semantics. Object-level operations account for the largest proportion at 4.38\%, followed by style-level (1.83\%), plot-level (0.65\%), scene-level (0.50\%), and other operations (0.44\%) of the full dataset.

We define chain length as the total number of video states in a trajectory, including the initial generation root and all subsequent edited states. Trajectories with 3, 4, and 5 states account for 61.5\%, 18.4\%, and 12.3\%, respectively, while 7.8\% contain more than 5 states. Thus, 92.2\% of the trajectories represent compact iterative workflows with 3--5 states, whereas the remainder captures longer creation trajectory.

\textbf{Root-Level Semantic Statistics.}
As shown in Figure~\ref{fig:four_images_row}, generation roots account for 17.8\% of all structured records. Within the generation subset, people-related content dominates at 13.68\%, followed by animals (2.33\%), transportation (0.49\%), scenery (0.43\%), and other categories (0.86\%) of the full dataset, respectively. Since the initial generations in multi-turn trajectories exhibit a similar semantic distribution, we report only the aggregate generation statistics.

Most videos are approximately 10 seconds long and contain around 300 frames at 30 FPS, while approximately 65.5\% exhibit a multi-shot structure. For prompt length, 52.2\% contain no more than 10 words, followed by 16.6\% with 11--25 words, 10.3\% with 26--50 words, 10.7\% with 51--100 words, and 10.3\% with more than 100 words.

\textbf{Edge-Level Operation Statistics.}
Within the single-turn editing subset, object-level operations account for 38.68\%, followed by style-level (20.09\%), scene-level (6.70\%), narrative and temporal (3.72\%), and other operations (5.21\%) of the full dataset, respectively. Generation roots and editing edges provide the initial states and state transitions that compose creation trajectories. Because the video-level attributes of the single-turn and multi-turn editing subsets closely match those of the generation subset, we omit their duplicate distributions.

\subsection{Natural Distribution and Annotation Reliability}

The distributions in Sora100K are naturally imbalanced, with people-related generation samples and several common editing operations accounting for relatively large proportions. Rather than constructing a conventionally balanced benchmark, Sora100K preserves the natural creation processes observed on an AI-Native video platform~\cite{openai2024sora} . The semantic categories of generation roots and editing operations of edges therefore reflect the creation intents observed during initial generation and successive refinement. We retain the corresponding annotations and metadata, allowing researchers to resample the data or construct balanced subsets for specific downstream tasks.

To assess the reliability of the VLM-based annotations used to characterize creation trajectories, we randomly sample generation roots and editing edges and compare the labels produced by Qwen2.5-VL-7B~\cite{bai2025qwen25vl} with human annotations under the same taxonomies. The automatic annotations achieve an accuracy of 98.19\%, a Cohen's $\kappa$ of 0.959, and a weighted F1 score of 98.26\%. These results support the reliability of the semantic and editing-operation annotations used to characterize creation trajectories. 

\begin{figure}[t]
    \includegraphics[width=\linewidth]{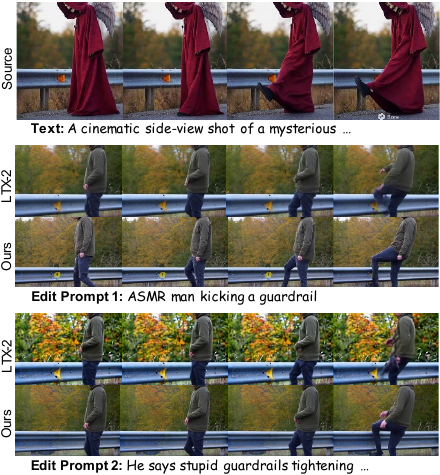}
    \vspace{-2em}
    \caption{Qualitative comparison on trajectory-level video editing. The Sora100K-adapted pipeline produces higher aesthetic quality while maintaining comparable subject consistency across successive editing turns.}
    \label{fig:trajectory}
\end{figure}

\section{Experiment}
We evaluate whether Sora100K provides effective supervision for trajectory-level video creation. Because current open-source models do not support direct training on complete multimodal trajectories, we adapt separate generation and editing components using trajectory roots and edges, then recombine them for successive-turn evaluation. We also evaluate each component independently to identify the sources of performance gains.

\subsection{Experiment Setting}
We construct an iterative pipeline with LTX-2-T2AV~\cite{hacohen2026ltx} as the generator and LTX-2-T2AV-IC~\cite{hacohen2026ltx} as the editor. Using LoRA~\cite{hu2022lora}, we adapt the generator on 3,000 generation roots and the editor on 3,000 editing edges. During evaluation, the generator produces the initial video from the root prompt, and the editor then applies the original instructions sequentially, using each output as the input to the next turn. We evaluate every resulting video state and compare the original and Sora100K-adapted pipelines. Detailed settings and more experiment samples are provided in the appendix.

\begin{table}[t]
\centering
\small
\setlength{\tabcolsep}{3.2pt}
\renewcommand{\arraystretch}{1.08}

\begin{tabular}{@{}llcccc@{}}
\toprule
\textbf{Turn}
& \textbf{Model}
& \textbf{$S_{\mathrm{eff}}\uparrow$}
& \textbf{$S_{\mathrm{Sub}}\uparrow$}
& \textbf{$S_{\mathrm{Aes}}\uparrow$}
& \textbf{Avg.$\uparrow$} \\
\midrule \midrule

\multirow{2}{*}{Edit 1}
& LTX-2
& 52.0
& \textbf{97.5}
& 41.5
& 63.7 \\
& Ours
& \textbf{54.0}
& 97.0
& \textbf{45.4}
& \textbf{65.5} \\
\midrule

\multirow{2}{*}{Edit 2}
& LTX-2
& \textbf{34.0}
& 97.0
& 41.5
& 57.5 \\
& Ours
& 32.0
& \textbf{97.7}
& \textbf{48.4}
& \textbf{59.4} \\
\midrule

\multirow{2}{*}{Average}
& LTX-2
& \textbf{43.0}
& 97.3
& 41.5
& 60.6 \\
& Ours
& \textbf{43.0}
& \textbf{97.4}
& \textbf{46.9}
& \textbf{62.4} \\
\bottomrule
\end{tabular}

\caption{Multi-turn video editing results over two successive editing turns. $S_{\mathrm{eff}}$ is a reference-based VLM score using the ground-truth edited video, while $S_{\mathrm{Sub}}$ and $S_{\mathrm{Aes}}$ are computed with VBench.}
\label{tab:trajectory_results}
\end{table}

\subsection{Experiment Results}

\hspace*{\parindent}\textbf{Trajectory-Level Video Creation Results.}
As shown in Figure~\ref{fig:trajectory}, we create a trajectory, and we evaluate each edited state using a VLM-based evaluator~\cite{bai2025qwen25vl} for editing effectiveness ($S_{\mathrm{eff}}$), subject consistency ($S_{\mathrm{Sub}}$), and aesthetic quality ($S_{\mathrm{Aes}}$), with their mean reported as the overall score. Table~\ref{tab:trajectory_results} shows that both pipelines degrade across turns. However, the Sora100K-adapted pipeline performs better, achieving a higher two-turn average of 62.4 versus 60.6.

The improvement primarily comes from aesthetic quality, whose two-turn average increases from 41.5 to 46.9. Subject consistency remains comparable, while editing effectiveness has the same two-turn average of 43.0 for both pipelines. These results show that Sora100K supervision improves the visual quality of video states throughout iterative creation, although successive editing remains challenging as errors accumulate across turns.

But the trajectory pipeline combines separate generation and editing models, the trajectory-level gains do not directly reveal which component benefits most from Sora100K supervision. We therefore conduct two controlled experiments that isolate the generation and editing stages, respectively, to identify the primary source of improvement.

\begin{figure}[t]
    \includegraphics[width=\linewidth]{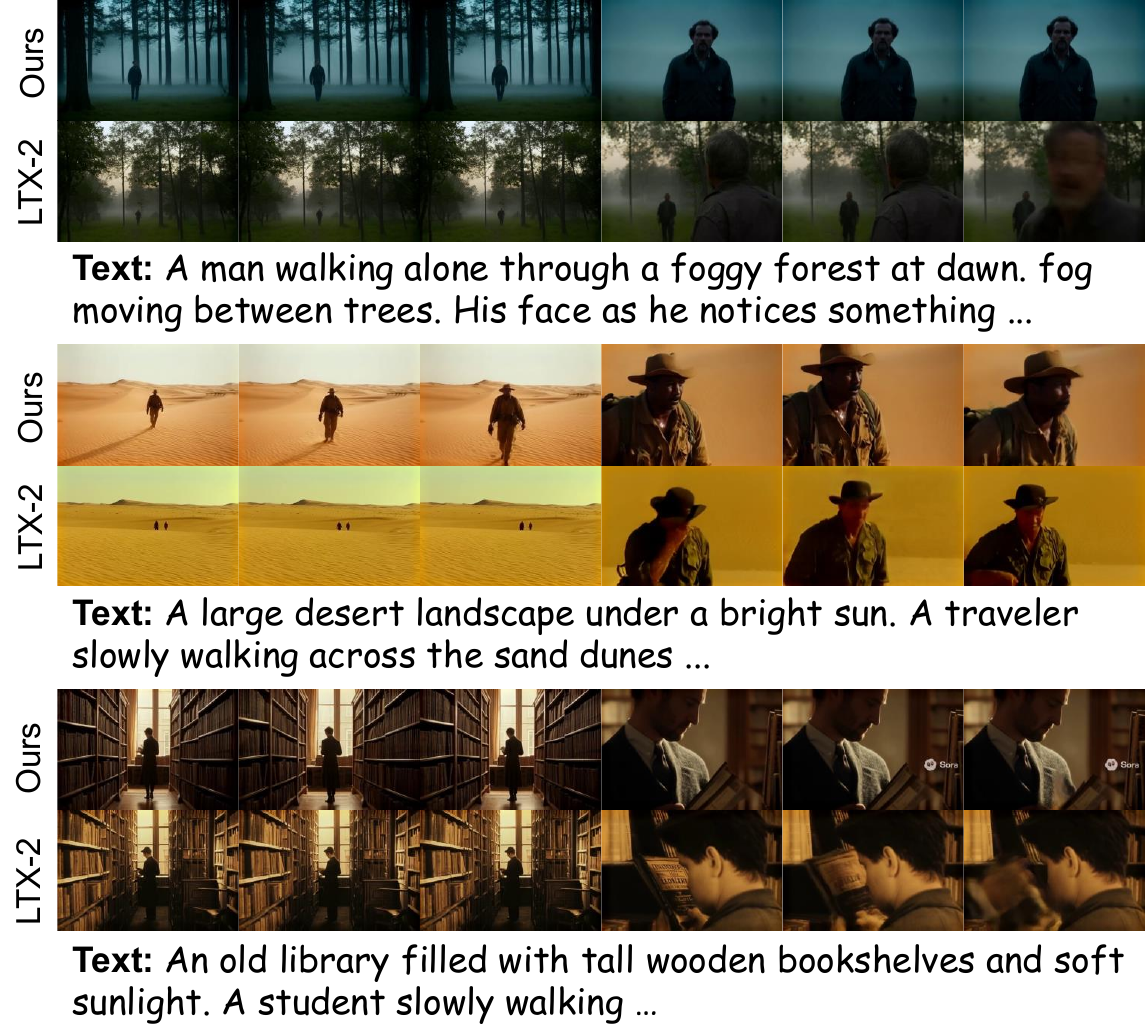}
    \caption{Qualitative comparison between LTX-2~\cite{hacohen2026ltx} and our model on video generation. Our model yields more consistent subjects, and fewer structural artifacts across shots.}
    \label{fig:video_analy}
\end{figure}

\begin{table*}[t]
\centering
\small
\setlength{\tabcolsep}{5pt}
\renewcommand{\arraystretch}{1.08}

\begin{tabular}{lcccccccc}
\toprule
\textbf{Model}
& \makecell[c]{\textbf{Subject}\\\textbf{Consistency}$\uparrow$}
& \makecell[c]{\textbf{Background}\\\textbf{Consistency}$\uparrow$}
& \makecell[c]{\textbf{Aesthetic}\\\textbf{Quality}$\uparrow$}
& \makecell[c]{\textbf{Imaging}\\\textbf{Quality}$\uparrow$}
& \makecell[c]{\textbf{Motion}\\\textbf{Smoothness}$\uparrow$}
& \textbf{ArcFace$\uparrow$}
& \textbf{Avg.$\uparrow$} \\
\midrule
\midrule
LTX-2(Base)
& 86.07
& 92.07
& 59.11
& 47.84
& 99.60
& 48.30
& 72.17 \\

Panda-70M
& 86.73
& 91.23
& 49.00
& 46.66
& \textbf{99.63}
& 49.23
& 70.41 \\
\midrule

Sora100K
& \textbf{86.90}
& \textbf{92.24}
& \textbf{62.98}
& \textbf{57.22}
& 99.56
& \textbf{59.70}
& \textbf{76.43} \\
\bottomrule
\end{tabular}

\caption{Quantitative comparison of the base LTX-2 model and its variants fine-tuned on 3,000 samples from Panda-70M and Sora100K, respectively. All fine-tuned variants use the same training configuration. Bold indicates the best result.}
\label{tab:ltx2}

\end{table*}

\begin{figure}[t]
    \includegraphics[width=\linewidth]{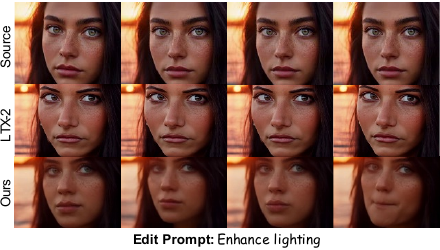}
    \vspace{-2em}
    \caption{Qualitative comparison between LTX-2~\cite{hacohen2026ltx} and our model on video editing. Our model achieves better aesthetic quality, editing effectiveness and stronger instruction following.}
    \label{fig:video_edit_analy}
\end{figure}

\begin{table}[t]
\centering
\small
\setlength{\tabcolsep}{3.5pt}
\renewcommand{\arraystretch}{1.08}
\begin{tabular}{lccccc}
\toprule
\textbf{Model} &
\textbf{$S_{\mathrm{inst}}\uparrow$} &
\textbf{$S_{\mathrm{eff}}\uparrow$} &
\textbf{$S_{\mathrm{Sub}}\uparrow$} &
\textbf{$S_{\mathrm{Aes}}\uparrow$} &
\textbf{Avg.$\uparrow$} \\
\midrule
\midrule
LTX-2
& 43.4 & 38.7 & \textbf{92.0} & 49.8 & 56.0 \\
Ours
& \textbf{47.9} & \textbf{44.2} & 91.7 & \textbf{53.8} & \textbf{59.4} \\
\bottomrule
\end{tabular}
\caption{Single-turn video editing results. The average is the arithmetic mean of the four reported metrics.}
\label{tab:single_turn_editing}
\end{table}

\textbf{Root-Level Video Generation Experiments.}
We compare the original LTX-2 generator with models adapted on Panda-70M~\cite{chen2024panda} and Sora100K using the same 200 held-out prompts. As shown in Table~\ref{tab:ltx2}, Sora100K achieves the best overall score of 76.43 and leads on five of the six metrics. Compared with the original model, it improves aesthetic quality from 59.11 to 62.98, imaging quality from 47.84 to 57.22, and ArcFace~\cite{deng2019arcface}  from 48.30 to 59.70. The substantial ArcFace gain indicates stronger identity consistency across shots, while motion smoothness remains nearly unchanged. Figure~\ref{fig:video_analy} further shows that the Sora100K-adapted model produces cleaner structures and more consistent subjects across shots. These results confirm that generation-root supervision provides a stronger initial state for trajectory construction.

\textbf{Edge-Level Video Editing Experiments.}
We next evaluate whether Sora100K improves individual state transitions along a trajectory. A VLM-based evaluator~\cite{bai2025qwen25vl} measures instruction following ($S_{\mathrm{inst}}$), editing effectiveness ($S_{\mathrm{eff}}$), subject consistency ($S_{\mathrm{Sub}}$), and aesthetic quality ($S_{\mathrm{Aes}}$). As shown in Table~\ref{tab:single_turn_editing}, adaptation improves $S_{\mathrm{inst}}$ from 43.4 to 47.9, $S_{\mathrm{eff}}$ from 38.7 to 44.2, and $S_{\mathrm{Aes}}$ from 49.8 to 53.8, increasing the average score from 56.0 to 59.4. Subject consistency remains nearly unchanged, decreasing slightly from 92.0 to 91.7. Figure~\ref{fig:video_edit_analy} shows corresponding improvements in editing strength and visual quality. These results demonstrate that editing-edge supervision improves the transitions between successive video states.

Generation roots improve the initial state, while editing edges strengthen subsequent transitions; however, performance still declines across turns.



\section{Conclusion}

We introduce Sora100K, a dataset that extends video data from isolated clips to structured audio-visual creation trajectories. It represents text-to-video generation, single-turn video editing, and multi-turn video editing subsets as roots, editing edges, and complete trajectories, while preserving source-to-edit lineage, intermediate video states. It also retains multi-shot structure and synchronized audio, more faithfully capturing key properties of modern AI-generated videos. Sora100K is constructed through systematic filtering, workflow reconstruction, structured annotation, and human validation, providing a curated resource for studying AI-Native video creation. Experiment results show that Sora100K provides effective supervision for video generation and single-turn editing, thereby enhancing performance in areas such as visual quality, cross-shot consistency, editing effectiveness and stronger instruction following.

\textbf{Limitations}.
First, suitable open-source models remain limited; many~\cite{team2026mova, zhang2026omninft} are derived from LTX-2 or incompatible with our input modalities. Second, creation trajectories remain beyond the capabilities of most open-source models, those are close-source.

\bibliography{aaai2027}


\clearpage
\appendix
\section{Appendix}
\label{app:appendix}

\begin{figure}[!b]
	\includegraphics[width=\linewidth]{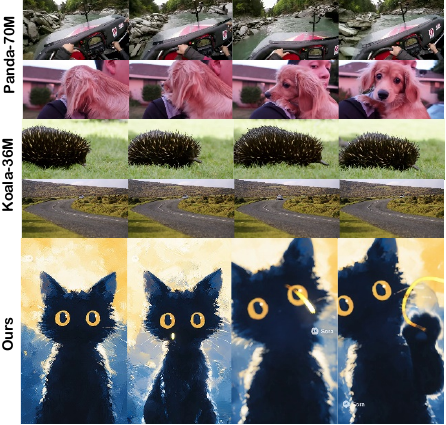}
	\vspace{-2em}
	\caption{Representative text-to-video samples from Panda-70M, Koala-36M, and Sora100K, illustrating their differences in shot organization and visual content.}
	\label{fig:trajectory1}
\end{figure}

\begin{figure}[!b]
	\includegraphics[width=\linewidth]{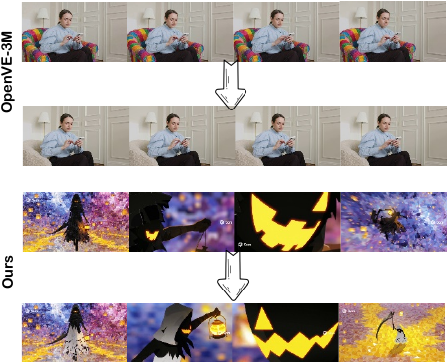}
	\vspace{-2em}
	\caption{Representative video editing samples from OpenVE-3M and Sora100K. The source and edited videos illustrate different types of content and style modifications.}
	\label{fig:trajectory2}
\end{figure}

\begin{figure*}[!t]
	\centering
	\includegraphics[width=\textwidth]{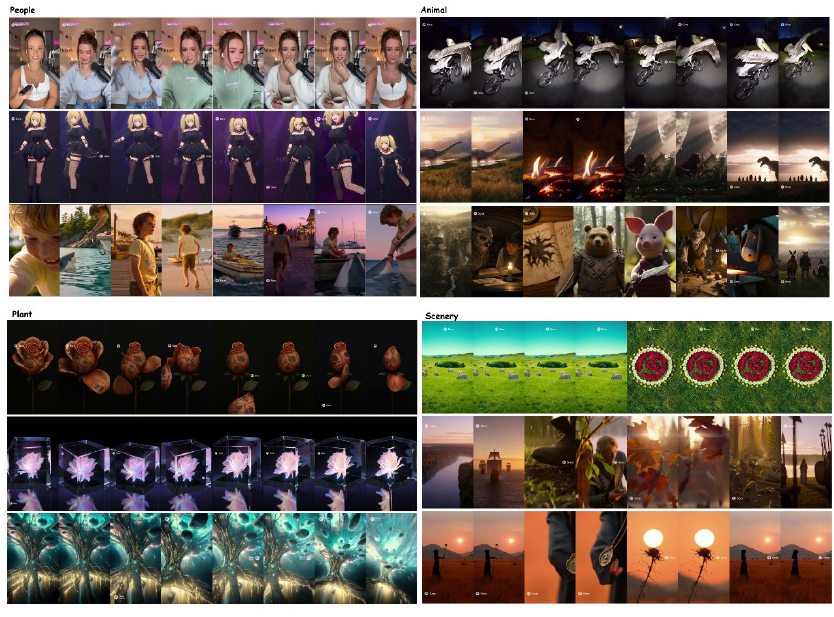}
	\caption{Representative samples from the Text-to-Video Generation
		subset across eight semantic categories: People, Animals, Architecture,
		Food \& Drink, Plants, Scenery, Science \& Technology, and Transportation.}
	\label{fig:show0}
\end{figure*}

\begin{figure*}[!t]
	\centering
	\includegraphics[width=\textwidth]{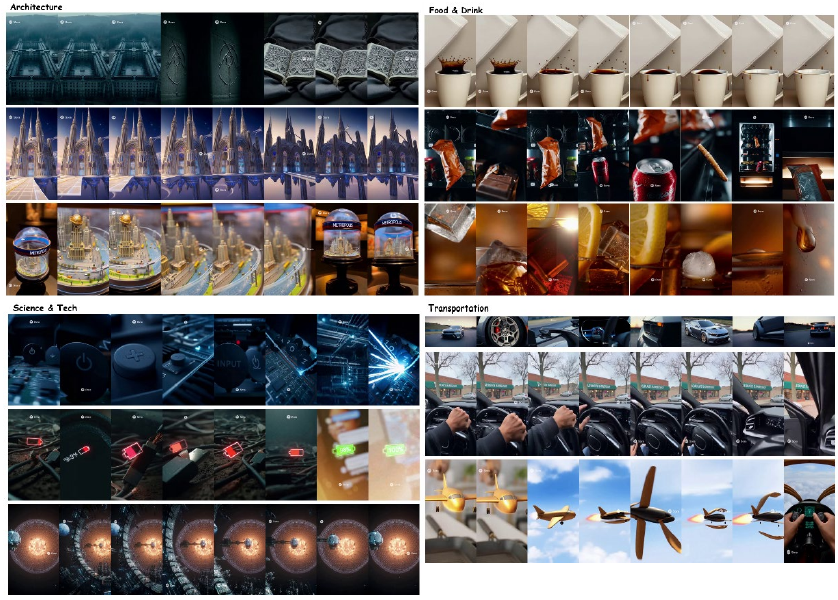}
	\caption{Representative samples from the Text-to-Video Generation
		subset across eight semantic categories: People, Animals, Architecture,
		Food \& Drink, Plants, Scenery, Science \& Technology, and Transportation.}
	\label{fig:show}
\end{figure*}

\begin{figure*}[!t]
	\centering
	\includegraphics[width=\textwidth]{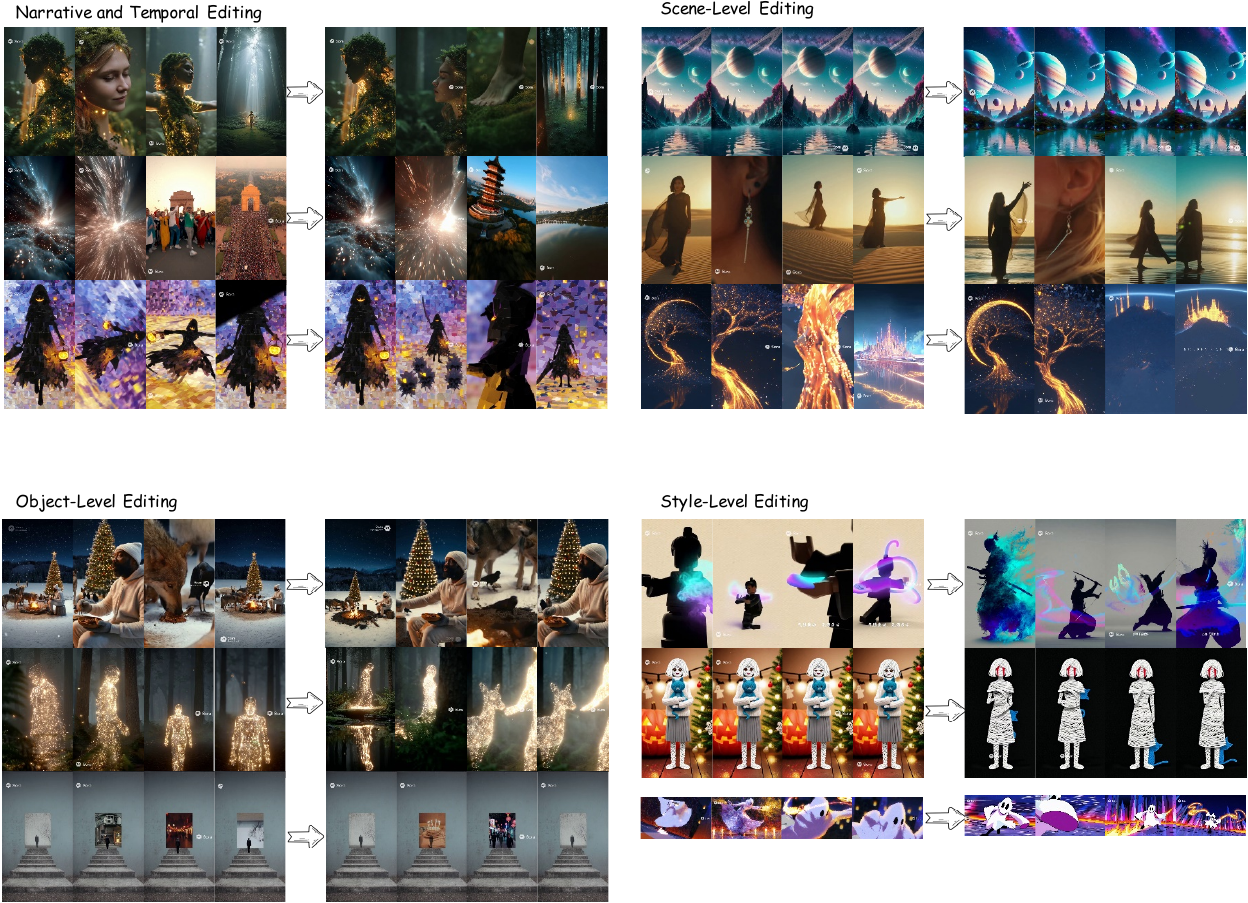}
	\caption{Representative samples from the Single-Turn Video Editing subset across four editing categories. Each example shows the source video on the left and the edited result on the right, covering object-level, style-level, scene-level, and narrative and temporal editing operations.}
	\label{fig:show1}
\end{figure*}
\begin{figure}[t]
	\centering
	\includegraphics[width=0.9\linewidth]{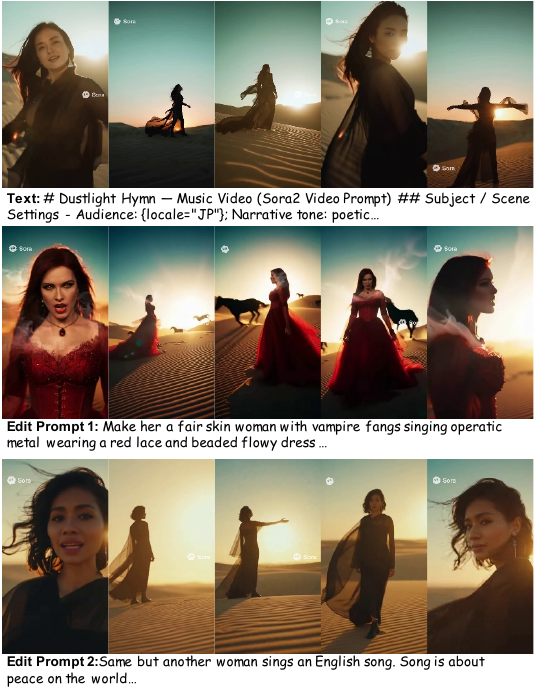}
	\caption{Representative multi-turn video editing trajectory.}
	\label{fig:trajectory15}
\end{figure}

\section{Motivation and Dataset Gap}
As shown in Figure~\ref{fig:trajectory1} -- Figure~\ref{fig:trajectory2}, we present a comparison of the proposed dataset with existing representative video datasets. Existing public datasets for video generation and editing remain primarily organized around clip-level, no-audio video samples, offering very limited explicit support for multi-shot structures and audio-visual video creation. Consequently, these datasets are currently insufficient for investigating long-range temporal consistency, cross-shot composition, and audiovisual coherence. 

Compared to existing video generation and editing datasets, Sora100K stands out for its more systematic coverage of key attributes such as long-term content, multi-shot structures, and audio-visual synchronization. These attributes are more aligned with current trends in video generation and the creative needs of real users. Therefore, this not only enhances the dataset's relevance to real-world scenarios but also provides a more supportive resource foundation for research on long-range dependency modeling, cross-shot organization, and audio-visual collaborative generation.

\begin{table*}[t]
	\centering
	\small
	\setlength{\tabcolsep}{3.2pt}
	\renewcommand{\arraystretch}{1.02}
	
	\begin{tabularx}{0.8\textwidth}{
			@{}
			>{\raggedright\arraybackslash}p{0.21\textwidth}
			>{\centering\arraybackslash}p{0.07\textwidth}
			>{\raggedright\arraybackslash}p{0.25\textwidth}
			>{\raggedright\arraybackslash}X
			@{}
		}
		\toprule
		\textbf{Field}
		& \textbf{Type}
		& \textbf{Description}
		& \textbf{Example} \\
		\midrule
		
		\multicolumn{4}{@{}l}{\textbf{Text-to-Video Generation}} \\
		
		\texttt{sample\_id}
		& string
		& Unique sample identifier
		& \texttt{s\_68dc...5177} \\
		
		\texttt{prompt}
		& string
		& Original generation prompt
		& Three guys getting lunch ... \\
		
		\texttt{description\_qwen}
		& string
		& Qwen2.5-VL-generated description
		& Three young men walk along a busy city street on a sunny day. \\
		
		\texttt{duration/frame/scene}
		& tuple
		& Duration, frame count, and shot count
		& \texttt{10/300/1} \\
		
		\texttt{category}
		& string
		& Generation category
		& \texttt{Real\_People} \\
		
		\midrule
		\multicolumn{4}{@{}l}{\textbf{Single-Turn Video Editing}} \\
		
		\texttt{sample/source/id}
		& string
		& Sample, source-video, and edited-video identifiers
		& \texttt{s\_68de...3765 / s\_68e1...ee3c} \\
		
		\texttt{edit\_instruction}
		& string
		& Original editing instruction
		& Change to a girl. \\
		
		\texttt{edit\_labels}
		& list
		& Editing category labels
		& \texttt{[object\_modification, style\_transfer]} \\
		
		\texttt{duration/frame/scene}
		& tuple
		& Duration, frame count, and shot count
		& \texttt{10/300/1} \\
		
		\midrule
		\multicolumn{4}{@{}l}{\textbf{Multi-Turn Video Editing}} \\
		
		\texttt{folder\_name}
		& string
		& Unique creation-trajectory identifier
		& \texttt{s\_68de...1d28} \\
		
		\texttt{chain\_length}
		& int
		& Number of videos in the trajectory
		& 3 \\
		
		\bottomrule
	\end{tabularx}
	
	\caption{Field definitions and representative examples for the three
		Sora100K subsets.}
	\label{tab:dataset_fields}
\end{table*}

\begin{table*}[t]
	\centering
	\small
	\setlength{\tabcolsep}{4pt}
	\renewcommand{\arraystretch}{1.04}
	
	\begin{tabularx}{0.8\textwidth}{
			@{}
			>{\raggedright\arraybackslash}p{0.20\textwidth}
			>{\raggedright\arraybackslash}X
			>{\raggedright\arraybackslash}X
			@{}
		}
		\toprule
		\textbf{Configuration}
		& \textbf{Text-to-Video Generation}
		& \textbf{Single-Turn Video Editing} \\
		\midrule
		
		Base model
		& LTX-2
		& LTX-2-T2AV-IC \\
		
		Training subset
		& Text-to-Video Generation
		& Single-Turn Video Editing \\
		
		Training samples
		& 3,000
		& 3,000 \\
		
		Input
		& Video and audio
		& Source video, target video, and audio \\
		
		Resolution
		& $512\times768$
		& $512\times768$ \\
		
		Number of frames
		& 121
		& 81 \\
		
		Fine-tuned component
		& DiT transformer
		& DiT transformer \\
		
		Adaptation method
		& LoRA
		& LoRA \\
		
		Learning rate
		& $1\times10^{-4}$
		& $1\times10^{-4}$ \\
		
		Epochs
		& 5
		& 5 \\
		
		LoRA rank
		& 32
		& 32 \\
		
		LoRA target modules
		& \texttt{to\_k}, \texttt{to\_q}, \texttt{to\_v},
		and \texttt{to\_out.0}
		& \texttt{to\_k}, \texttt{to\_q}, \texttt{to\_v},
		and \texttt{to\_out.0} \\
		
		Gradient checkpointing
		& Enabled
		& Enabled \\
		
		Hardware
		& One NVIDIA H100 GPU
		& One NVIDIA H100 GPU \\
		
		Peak GPU memory
		& Approximately 40\,GB
		& Approximately 40\,GB \\
		
		\bottomrule
	\end{tabularx}
	
	\caption{Training configurations for LoRA adaptation on the
		Text-to-Video Generation and Single-Turn Video Editing subsets.}
	\label{tab:ltx2_training_configurations}
\end{table*}

\section{Dataset Samples}

\subsection{Text-to-Video Generation Subset}
In this section, we will present more samples from our dataset, with each sample displayed according to our classification categories.
For Text-to-Video Generation, we organize all samples into eight semantic categories: \textbf{People}, \textbf{Animal}, \textbf{Plant}, \textbf{Food \& Drink}, \textbf{Transportation}, \textbf{Scenery}, \textbf{Architecture}, and \textbf{Science \& Tech}. 
Each generation video is assigned a primary label based on its main narrative theme.
Each samples as shown in Figure~\ref{fig:show0} -- Figure~\ref{fig:show}.

	


\noindent\textbf{People.}
This category includes videos in which one or more human figures constitute the primary visual subject, covering portraits, social interactions, performances, sports, professional activities, and everyday behaviors. A video is assigned to this category when people remain the central focus, even if animals, vehicles, or environmental elements are also present. This category contains 14,169 samples.

\noindent\textbf{Animal.}
This category includes videos in which animals constitute the primary visual subject, including pets, wildlife, birds, insects, marine animals, and anthropomorphic animals. This category contains 2,417 samples.

\noindent\textbf{Transportation.}
This category includes videos primarily depicting vehicles or transportation-related activities, such as cars, trains, aircraft, ships, bicycles, motorcycles, and other forms of transportation. This category contains 511 samples.

\noindent\textbf{Scenery.}
This category includes videos primarily depicting natural environments or broad landscapes without a dominant human, animal, or human-made subject. Representative content includes mountains, forests, beaches, rivers, deserts, skies, and weather-related scenes. This category contains 446 samples.

\noindent\textbf{Food \& Drink.}
This category includes videos centered on food, beverages,ingredients, cooking processes, food preparation, plating, or eating and drinking scenes. This category contains 380 samples.

\noindent\textbf{Architecture.}
This category includes videos in which buildings or human-made spaces constitute the primary visual subject, including houses, palaces, temples, bridges, urban structures, interior spaces, and other architectural environments. This category contains 237 samples.

\noindent\textbf{Science \& Technology.}
This category includes videos primarily depicting scientific or technological subjects, such as robots, electronic devices, computer interfaces, laboratories, spacecraft, industrial equipment, and futuristic technologies. This category contains 234 samples.

\noindent\textbf{Plants.}
This category includes videos in which plants themselves constitute the primary visual subject, including flowers, trees, crops, potted plants, and plant-growth processes. Plants appearing only as part of a broader natural environment are instead categorized as \textit{Scenery}. This category contains 44 samples.

\subsection{Single-Turn Video Editing Subset}
We further analyze the Single-Turn Video Editing subset along four dimensions: editing category, video duration, frame count, and number of shots. Based on the automatic annotations produced by Qwen2.5-VL, object-level editing accounts for the largest proportion at 52\%, followed by style-level editing at 27\%. Temporal and scene-level editing account for 9\% and 5\%, respectively, while the remaining 7\% consists of operations that are not fully covered by the predefined taxonomy. These distributions show that Sora100K covers common video editing operations while retaining diverse and open-ended editing requests. Each samples as shown in Figure~\ref{fig:show1}.

\noindent\textbf{Object-Level Editing.}
This category includes localized modifications to foreground objects, people, animals, or other visible entities. Typical operations include adding, removing, replacing, duplicating, resizing, repositioning, recoloring, or modifying the attributes of an entity while largely preserving the surrounding environment and overall visual style. Object-level editing accounts for 52\% of the Single-Turn Video Editing subset.

\noindent\textbf{Style-Level Editing.}
This category includes modifications to the visual appearance of a video without fundamentally changing its primary subjects or scene structure. Typical operations include changing the artistic style, color palette, lighting, texture, rendering mode, or overall cinematic tone. Style-level editing accounts for 27\% of the subset.

\noindent\textbf{Temporal Editing.}
This category includes modifications to actions, motion dynamics, or the temporal organization of a video. Typical operations include changing an action, motion speed, movement direction, motion  trajectory, event order, duration, or temporal continuity. Temporal editing accounts for 9\% of the subset.

\noindent\textbf{Scene-Level Editing.}
This category includes modifications to the broader environment or spatial composition of a video. Typical operations include changing the background, location, weather, time of day, scene layout, viewpoint, or camera framing while retaining the main subject when appropriate. Scene-level editing accounts for 5\% of the subset.

\noindent\textbf{Other Editing.}
This category includes uncommon, compound, or ambiguous editing operations that cannot be reliably assigned to the predefined object-, style-, temporal-, or scene-level categories. It also covers instructions that simultaneously modify multiple aspects without a clearly dominant editing intent. This category accounts for the remaining 7\% of the subset.

\subsection{Multi-Turn Video Editing Subset}
The Multi-Turn Video Editing subset captures iterative creation processes in which each trajectory starts from an initial generated video and evolves through multiple successive editing turns. Unlike the Single-Turn Video Editing subset, which represents individual source--instruction--target transitions, this subset preserves the ordered sequence of intermediate video states and editing instructions. Each samples as shown in Figure~\ref{fig:trajectory15}.


\section{Dataset Annotation and Statistics}
\label{app:dataset_statistics}

This section presents the annotation fields and detailed statistics of the three subsets in Sora100K: Text-to-Video Generation, Single-Turn Video Editing, and Multi-Turn Video Editing. We first describe the fields provided for each subset and then report their composition, semantic categories, editing categories, video attributes, prompt lengths, and editing chain lengths.

Table~\ref{tab:dataset_fields} summarizes the fields provided for the three subsets of Sora100K. The Text-to-Video Generation subset records the original prompt, supplementary description, video attributes, and semantic category. The Single-Turn Video Editing subset records each source--instruction--target transition together with its editing labels and video statistics. The Multi-Turn Video Editing subset further organizes successive video states into editing chains and records the length of each chain.

\subsection{Annotation Fields}
\label{app:annotation_fields}

\noindent\textbf{Text-to-Video Generation Subset.}
Each sample in the Text-to-Video Generation subset contains seven fields. The \texttt{sample\_id} field provides a unique sample identifier, while \texttt{prompt} stores the original generation
prompt. The \texttt{description\_qwen} field contains a supplementary video description generated by Qwen2.5-VL. The video attributes are recorded by \texttt{duration\_sec}, \texttt{frame\_count}, and \texttt{scene\_count}, which represent the video duration in seconds, the number of frames, and the number of scenes or shots, respectively. The \texttt{category} field stores the generation category label of the sample.

\noindent\textbf{Single-Turn Video Editing Subset.}
Each sample in the Single-Turn Video Editing subset represents one editing transition. The \texttt{sample\_id} field identifies the sample, while \texttt{source\_video\_id} and \texttt{edited\_video\_id} identify the source and edited videos, respectively. The \texttt{edit\_instruction} field stores the original editing instruction. The \texttt{edit\_labels} field records the corresponding editing category labels, such as object modification and style transfer. The subset also records the video duration, frame count, and scene or shot count of each sample.

\noindent\textbf{Multi-Turn Video Editing Subset.}
The Multi-Turn Video Editing subset organizes successively edited videos into ordered editing chains. Each chain is identified by the \texttt{folder\_name} field. The \texttt{chain\_length} field records the number of videos contained in the chain.

\subsection{Overall Dataset Composition}
\label{app:overall_composition}

Sora100K contains 103,439 entries across its three subsets. The Text-to-Video Generation subset contains 18,451 entries and accounts for 17.8\% of the dataset. The Single-Turn Video Editing subset contains 76,964 entries, representing the largest proportion at 74.4\%. The remaining 8,024 entries belong to the Multi-Turn Video Editing subset, accounting for 7.8\%.

\subsection{Text-to-Video Generation Statistics}
\label{app:generation_statistics}

\noindent\textbf{Semantic Category Distribution.}
The semantic distribution of the Text-to-Video Generation subset is dominated by people-related content. The \textit{People} category accounts for 76.85\% of the subset, followed by \textit{Animal} at 13.11\%. \textit{Transportation} and \textit{Scenery} account for 2.77\% and 2.42\%, respectively. The remaining categories are grouped as \textit{Other}, which accounts for 4.85\%.

\noindent\textbf{Prompt Length Distribution.}
Generation prompts are divided into five length ranges. Prompts containing no more than 10 words account for 52.20\% of thesubset. Prompts containing 11--25 words account for 16.55\%, while those containing 26--50 and 51--100 words account for 10.28\% and 10.65\%, respectively. Prompts longer than 100 words account for the remaining 10.32\%.

\noindent\textbf{Video Duration.}
Most videos have durations close to 10 seconds. Videos lasting between 9.5 and 10.5 seconds account for 73.48\% of the dataset, while videos lasting between 10.5 and 15 seconds account for 23.84\%. Videos no longer than 9.5 seconds account for 2.57\%, and only 0.11\% are longer than 15 seconds.

\noindent\textbf{Scene Count.}
The scene-count distribution is reported over five ranges: 0, 1--3, 4--6, 7--10, and more than 10. Videos containing 4--6 or 7--10 scenes constitute the largest groups, while videos containing more than 10 scenes account for the smallest proportion.

\subsection{Single-Turn Video Editing Statistics}
\label{app:single_turn_statistics}

The Single-Turn Video Editing subset covers five major groups of editing operations. Object-level editing accounts for the largest proportion at 52\%, followed by style-level editing at 27\%. Scene-level editing accounts for 9\%, while narrative and temporal editing accounts for 5\%. The remaining 7\% is categorized as \textit{Other}.

\noindent\textbf{Object-Level Editing.}
Object-level editing modifies visible entities or their attributes within a video. This group accounts for 52\% of the Single-Turn Video Editing subset.

\noindent\textbf{Style-Level Editing.}
Style-level editing changes the visual appearance or presentation style of a video. This group accounts for 27\% of the subset.

\noindent\textbf{Scene-Level Editing.}
Scene-level editing changes the scene or environmental content of a video. This group accounts for 9\% of the subset.

\noindent\textbf{Narrative and Temporal Editing.}
Narrative and temporal editing changes the narrative content or temporal development of a video. This group accounts for 5\% of the subset.

\subsection{Multi-Turn Video Editing Statistics}
\label{app:multi_turn_statistics}

The Multi-Turn Video Editing subset contains ordered video chains formed through successive editing operations. We use chain length, defined as the number of videos contained in an editing chain, to describe the structural distribution.

\section{Experiment}

\subsection{Experiment Setting}
\noindent\textbf{Implementation Details.}
For the text-to-video generation experiment, we randomly sample 3,000 videos from the Text-to-Video Generation subset and adapt the base LTX-2 model using LoRA. Each training sample retains its audio input and is resized to a resolution of $512\times768$ with 121 frames. We train the model for five epochs using a learning rate of $1\times10^{-4}$. The LoRA rank is set to 32, and LoRA adapters are applied to the \texttt{to\_k}, \texttt{to\_q}, \texttt{to\_v}, and \texttt{to\_out.0} modules. Gradient checkpointing is enabled to reduce GPU memory consumption. The experiment is conducted on a single NVIDIA H100 GPU, with approximately 40\,GB of peak GPU memory usage. The detailed configuration is summarized in Table~\ref{tab:ltx2_training_configurations}.
We follow the official LTX-2 training workflow and divide the adaptation process into data preprocessing and LoRA fine-tuning. For the video editing experiments, we randomly sample 3,000 samples from the Single-Turn Video Editing subset and fine-tune LTX-2-T2AV-IC. Each training sample contains the edited video, input audio, and an in-context source video, together with the frame rate and context-downsampling information. All videos are resized to $512\times768$ and represented by 81 frames.
During preprocessing, we load the text-encoder post-processing modules, video VAE encoder, audio VAE encoder, and Gemma-3-12B weights to precompute the training representations. We then fine-tune the DiT transformer for five epochs using a learning rate of $1\times10^{-4}$. LoRA is applied to the \texttt{to\_k}, \texttt{to\_q}, \texttt{to\_v}, and \texttt{to\_out.0} modules with a rank of 32. Gradient checkpointing is enabled throughout training. All experiments are conducted on a single NVIDIA H100 GPU, with approximately 40\,GB of peak GPU memory usage. The detailed configuration is summarized in Table~\ref{tab:ltx2_training_configurations}.

\subsection{Experiment Result.}
\begin{figure*}[!t]
	\centering
	\includegraphics[width=\textwidth]{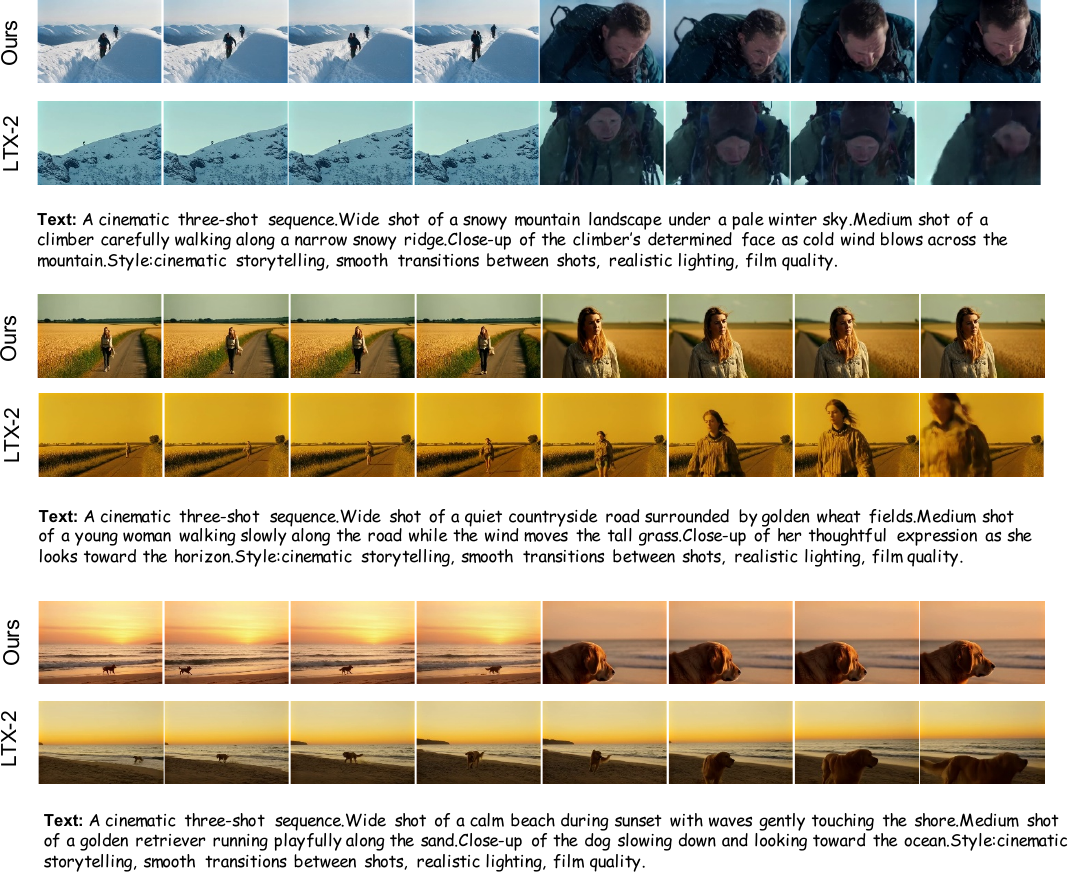}
	\caption{Qualitative comparison between the Sora100K-adapted model and LTX-2 on text-to-video generation. Each example presents sampled frames from the two models under the same prompt, illustrating their differences in visual quality, subject preservation, and consistency across shots. Red boxes highlight representative subject-consistency differences.}
	\label{fig:show4}
\end{figure*}

\begin{figure}[!t]
	\centering
	\includegraphics[
	width=\columnwidth,
	keepaspectratio
	]{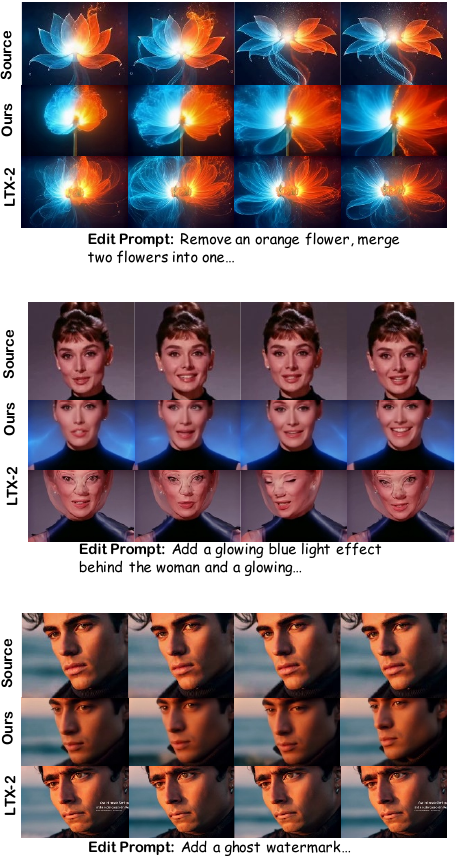}
	\caption{Qualitative comparison between the Sora100K-adapted model and LTX-2 on single-turn video editing. The examples cover object removal and merging, lighting-effect addition, and watermark insertion.}
	\label{fig:show5}
\end{figure}

\begin{figure}[!t]
	\includegraphics[width=0.5\textwidth]{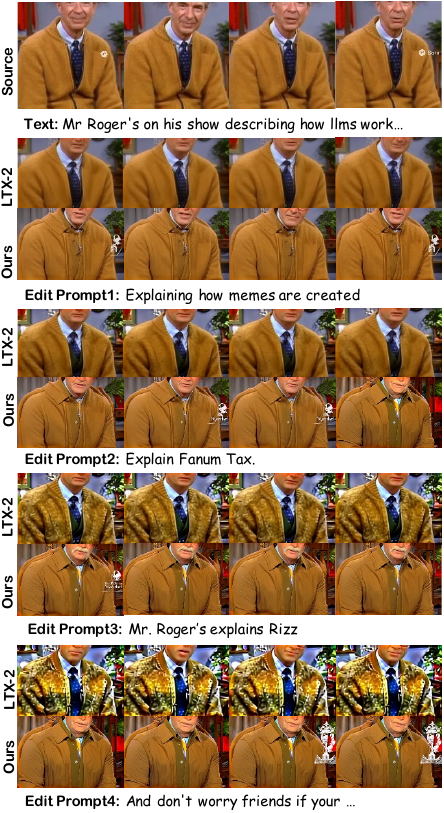}
	\caption{Qualitative comparison of multi-turn video editing between LTX-2 and our model. The models sequentially modify the source video according to four editing instructions. Our model better preserves character appearance and scene structure while accumulating changes across turns.}
	\label{fig:show7}
\end{figure}

\begin{figure}[t]
	\includegraphics[width=\linewidth]{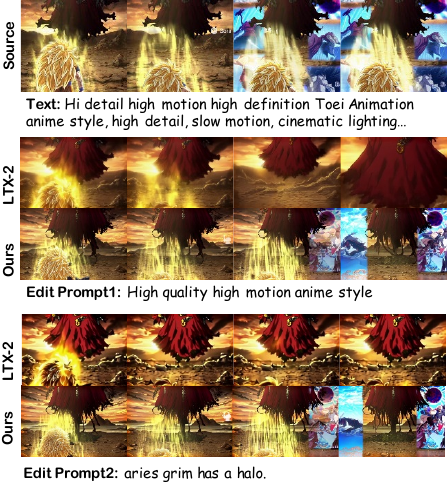}
	\vspace{-2em}
	\caption{Qualitative comparison of multi-turn video editing between LTX-2 and our model. Starting from the same source video, the models sequentially apply a high-motion anime style and add a halo to the character. Our model better preserves the visual content and previous editing effects across successive turns.}
	\label{fig:show6}
\end{figure}

As shown in Figure~\ref{fig:show4} -- Figure~\ref{fig:show6}, we provide additional qualitative results of multi-turn video creation trajectories. 

\end{document}